\documentclass[sigconf, nonacm]{acmart}
\AtBeginDocument{%
  }
\usepackage[table]{xcolor}
\usepackage{makecell}
\usepackage{microtype}
\usepackage{booktabs}
\usepackage{array}
\usepackage{subcaption} % 处理子图的关键包
\usepackage{graphicx}   % 处理图片的包
\usepackage{algorithm}
\usepackage{algorithmic}
\usepackage{amsmath}
\usepackage{url}
\usepackage{hyperref}
\usepackage{multirow}
\renewcommand{\arraystretch}{1.08}
\newcolumntype{L}[1]{>{\raggedright\arraybackslash}p{#1}}

\usepackage{xspace}
\newcommand{\STREAM}{\textit{STQuant}\xspace}

\usepackage{enumitem}
\setlist{topsep=2pt,itemsep=0pt,parsep=0pt,partopsep=0pt}

\setcopyright{acmlicensed}
\copyrightyear{2026}
\acmYear{2026}
\acmDOI{XXXXXXX.XXXXXXX}
\acmConference[Conference acronym 'XX]{Make sure to enter the correct
  conference title from your rights confirmation email}{
  2026}{Woodstock, NY}
\acmISBN{978-1-4503-XXXX-X/2018/06}

\begin{document}

%%
%% The "title" command has an optional parameter,
%% allowing the author to define a "short title" to be used in page headers.
\title{
% STREAM: Spatio-Temporal Regulated Adaptive Momentum Quantization for Memory-Efficient Large Multimedia Model Training
% A Spatio-Temporal Framework with Regulated Adaptive Momentum for Quantization in Large Multimodal Model Training
STQuant: Spatio-Temporal Adaptive Framework for Optimizer Quantization
in Large Multimodal Model Training
}

%%
%% The "author" command and its associated commands are used to define
%% the authors and their affiliations.
%% Of note is the shared affiliation of the first two authors, and the
%% "authornote" and "authornotemark" commands
%% used to denote shared contribution to the research.

%%
%% By default, the full list of authors will be used in the page
%% headers. Often, this list is too long, and will overlap
%% other information printed in the page headers. This command allows
%% the author to define a more concise list
%% of authors' names for this purpose.
% \renewcommand{\shortauthors}{Trovato et al.}

\author{Minglu Liu}
\affiliation{%
  \institution{Xidian University}
  \city{Xian}
  \state{Shaanxi}
  \country{China}
}
\email{25071213205@stu.xidian.edu.cn}

\author{Cunchen Hu}
\authornote{Cunchen Hu and Liangliang Xu are the corresponding authors.}
\affiliation{%
  \institution{China Telecom Cloud Computing Research Institute}
  % \city{Beijing}
  \state{Beijing}
  \country{China}
}
\email{hucunchen21@mails.ucas.ac.cn}

\author{Liangliang Xu}
\authornotemark[1]
\affiliation{%
  \institution{Xidian University}
  \city{Xian}
  \state{Shaanxi}
  \country{China}
}
\email{llxu@mail.ustc.edu.cn}

\author{Fengming Tang}
\affiliation{%
  \institution{Xidian University}
  \city{Xian}
  \state{Shaanxi}
  \country{China}
}
\email{fm.tang@stu.xidian.edu.cn}

\author{Ruijia Wang}
\affiliation{%
  \institution{China Telecom Cloud Computing Research Institute}
  % \city{Beijing}
  \state{Beijing}
  \country{China}
}
\email{wangrj12@chinatelecom.cn}

\author{Fu Yu}
\affiliation{%
  \institution{China Telecom Cloud Computing Research Institute}
  % \city{Beijing}
  \state{Beijing}
  \country{China}
}
\email{yufubupt@163.com}

%%
%% The abstract is a short summary of the work to be presented in the
%% article.
\begin{abstract}

Quantization is an effective way to reduce the memory cost of large-scale model training. However, most existing methods adopt fixed-precision policies, which ignore the fact that optimizer-state distributions vary significantly across layers and training steps. Such uniform designs often introduce noticeable accuracy degradation. To move beyond fixed quantization, we propose \STREAM, a distributed training framework that reduces the memory footprint of optimizer states via dynamic precision allocation across layers, state variables, and training steps, while maintaining model quality. Naively applying dynamic quantization during training is challenging for two reasons. First, optimizer states are numerically sensitive, and quantization noise can destabilize quality. Second, jointly considering multiple states and layers induces a large combinatorial search space. \STREAM addresses these challenges with two key techniques: 1) a provably near-optimal factor selection strategy that accurately identifies the most influential factors for precision adaptation. 2) a dynamic transition decision algorithm that reduces the search cost from exponential to linear complexity. Experiments on GPT-2 and ViT show that \STREAM reduces optimizer-state memory by 84.4\%, achieving an average bit-width of as low as 5.1 bits, compared with existing solutions. Moreover, \STREAM incurs only $O(N/K)$ computational overhead and requires $O(1)$ extra space.

\end{abstract}

%%
%% The code below is generated by the tool at http://dl.acm.org/ccs.cfm.
%% Please copy and paste the code instead of the example below.
%%


%% Keywords. The author(s) should pick words that accurately describe
%% the work being presented. Separate the keywords with commas.
%% A "teaser" image appears between the author and affiliation
%% information and the body of the document, and typically spans the
%% page.

\received{20 February 2007}
\received[revised]{12 March 2009}
\received[accepted]{5 June 2009}

%%
%% This command processes the author and affiliation and title
%% information and builds the first part of the formatted document.

\maketitle
\section{Introduction}
\begin{figure*}[t] % 注意这里的星号，它代表横跨双栏；[t] 代表放在页面顶部
  \centering
  \includegraphics[scale=0.28]{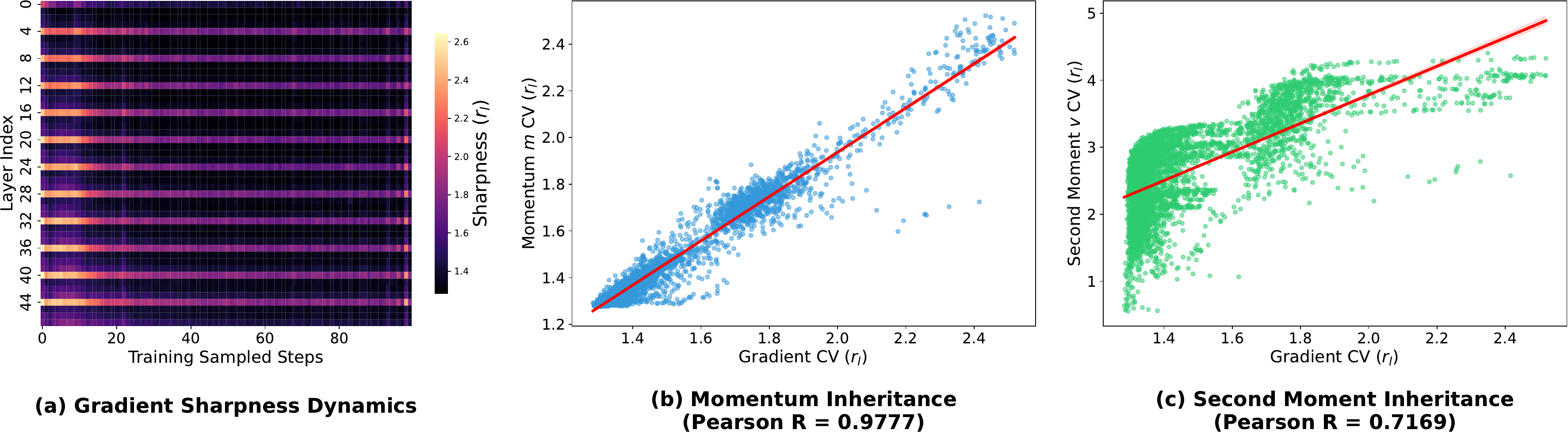} % width=\textwidth 确保图片撑满左右两栏的宽度
  \caption{\textbf{Spatiotemporal evolution and correlation analysis of gradients and Adam optimizer states.} 
(a) Sharpness of the gradient distribution (CV) across 48 core weight layers in 12 Transformer blocks, including QKV projections, attention projections, MLP expansion layers, and MLP compression layers, sampled every 50 training steps over 5000 total steps; 
(b) Correlation between gradients and the coefficient of variation (CV) of the first-order moment $m$, showing a very strong linear relationship (Pearson $R = 0.9777$); 
(c) Correlation between gradients and the second-order moment $v$, where, despite the squaring operation introducing nonlinear mapping and noise amplification, the correlation remains strong ($R = 0.7169$).}
  \label{fig:motivation}
\end{figure*}
As Large Foundation Models are widely applied in fields such as natural language processing \cite{zhao2023survey}, image generation \cite{ramesh2021zero}, and code generation \cite{roziere2023code}, the demand for model quality has increased, leading to explosive growth in parameter scale \cite{achiam2023gpt}.
Simultaneously, the storage requirements for model parameters and high-precision optimizer states have substantially increased memory consumption, becoming a major bottleneck for large-scale training. 
For example, in typical BF16 training, Adam optimizer states, consisting of first-order moment $m$ and second-order moment $v$, usually occupy $4\times$ to $6\times$ more memory than the model weights \cite{rajbhandari2020zero}. With the adoption of FP8 and lower-bit quantization for model weights, this ratio further increases to $8\times$ to $12\times$ \cite{zhao2024galore}. However, optimizer states are critical to training stability and model accuracy because they are extremely sensitive to quantization errors, where even minor discrepancies can lead to gradient explosion or loss divergence. Therefore, designing high-compression and low-precision loss quantization strategies for optimizer states is crucial for improving training efficiency and scalability.

% \ruijia{[Personal comments: The current narrative order feels slightly misaligned with the final "high-compression + low-precision loss" conclusion. Specifically, introducing the extreme sensitivity of optimizer states first creates an immediate intuition that we shouldn't touch or compress them, making the sudden pivot to memory bottlenecks feel abrupt. A much more compelling and natural logical flow would be: Model training scales up -> Optimizer states occupy huge memory -> We want to compress them, BUT they are highly sensitive -> Therefore, we need strategies that are both high-compression and lossless.]}

% Consequently, the issue of stability during the training process \cite{zhang2022opt} has become more prominent. In model training, optimizer states are critical factors affecting model precision; they are extremely sensitive to errors, where even minor discrepancies can lead to gradient explosion or loss divergence.

In recent years, a large body of work has explored optimizer-state quantization. For example, 8-bit Adam~\cite{dettmers20218} and FP8~\cite{micikevicius2022fp8} primarily rely on static uniform quantization. AnyPrecision~\cite{park2024any} introduces mixed-precision quantization for greater flexibility. In addition, Lion~\cite{chen2023symbolic} and Adam-mini~\cite{zhang2024adam} improve efficiency through lightweight optimizer designs.
However, optimizer-state quantization still faces limitations in training dynamics and search complexity.
Static quantization schemes~\cite{dettmers20218, micikevicius2022fp8} fail to capture the spatiotemporal non-stationarity of gradients during training, while fixed bit-width allocation or static block structures limit adaptability to training dynamics. Search-based mixed-precision methods~\cite{park2024any} suffer from exponentially growing search spaces and parameter coupling, making them difficult to apply to large-scale pretraining. 
% \ruijia{[Personal comments: 1. HAWQ-V2 is an inference-side weight quantization method, not for dynamic optimizer states. 2. Lion and Adam-mini are lightweight optimizers, not quantization -> yet the "However" clause only attacks quantization flaws, leaving these citations disconnected. ]}

% {\color{brown}In recent years, a large body of work has explored optimizer-state quantization. For example, 8-bit Adam~\cite{dettmers20218} and FP8-based low-precision training~\cite{micikevicius2022fp8} primarily rely on static uniform quantization, while methods such as AnyPrecision~\cite{park2024any} introduce more flexible precision allocation strategies. However, existing optimizer-state quantization methods still face challenges in capturing training dynamics and controlling search complexity. Static quantization schemes fail to model the spatiotemporal non-stationarity of gradients and optimizer statistics during training, while fixed bit-width allocation or static block structures limit adaptability to evolving optimization trajectories. Orthogonal to direct quantization, lightweight optimizers such as Lion~\cite{chen2023symbolic} and Adam-mini~\cite{zhang2024adam} improve efficiency through optimizer redesign. While effective in reducing optimizer overhead, these methods typically simplify Adam’s adaptive state modeling, which may limit their ability to fully capture evolving gradient statistics under non-stationary training dynamics. }

We observe that the sensitivity of optimizer states to quantization errors depends on the distribution of gradients during training.
% posing significant challenges for the design of quantization strategies. 
As shown in Figure.~\ref{fig:motivation}(a), gradients exhibit large fluctuations in the early stages of training (training sampled steps $<20$), which gradually stabilize in later stages (training sampled steps $>20$). Moreover, we also observe periodic horizontal stripes across layers, indicating that different layers exhibit varying tolerance to quantization errors. To quantitatively assess whether optimizer states inherit the spatiotemporal characteristics of gradients, we compute the \footnote{The Pearson correlation coefficient measures the linear correlation between two variables and is defined as the covariance normalized by the product of their standard deviations.}{Pearson} correlation coefficients between the gradients and the first-order moment $m$ and second-order moment $v$ of the optimizer states (Figure.~\ref{fig:motivation}(b) and \ref{fig:motivation}(c)). We find a strong spatiotemporal correlation between optimizer states and gradients, suggesting that the optimizer states are not randomly distributed but are determined by the gradients’ intrinsic physical properties. Consequently, the optimal optimizer quantization requires considering the dynamism across training steps (temporal adaptivity) and the sensitivity to layer-specific structures (layer-wise adaptivity). 
To address the above challenges, we propose \STREAM, a general \textbf{\underline{s}}patio-\textbf{\underline{t}}emporal adaptive framework for optimizer \textbf{\underline{quant}}ization in model training, designed to reduce memory consumption while preserving training stability with low overhead.
The key idea behind \STREAM is to capture optimizer-state heterogeneity along both temporal and spatial dimensions.
(1) temporal: Low-bit quantization can be unstable early in training and redundant later. \STREAM introduces an adaptive, stages-aware quantization strategy inspired by simulated annealing and alleviates quantization-induced degradation on the convergence trajectory.
%
% inspired by simulated annealing, \STREAM introduces a temporal factor that enables high-precision buffering in the early training stage, where random initialization often induces instability. This design mitigates the impact of quantization noise on the convergence trajectory.
%
(2) spatial: To account for the heterogeneous sensitivity of different layers to accumulated errors, \STREAM adopts a heuristic scoring mechanism that reframe complex global optimization as gradient-statistics-based bit-width selection, enabling real-time identification and protection of critical layers.
% To account for the heterogeneous sensitivity of model layers to accumulated errors, \STREAM adopts a heuristic scoring strategy that converts complex global optimization into gradient-statistics-based bit-width selection, enabling real-time identification and protection of critical layers.
Specifically, \STREAM comprises three engines: 1) a \textit{score engine} for computing gradient statistics; 2) a \textit{distributed engine} for synchronizing them across GPUs and determining layer-wise bit-widths; and 3) a \textit{quantization engine} for applying block-wise quantization to optimizer states.
In summary, we make the following contributions. 
% \ruijia{[Personal comments: 1. Moving the empirical observations about the characteristics of gradients to this paragraph may be a better structural pivot. Its current placement lacks a logical connection to the surrounding text, whereas moving it here serves perfectly as the explicit motivation for our spatial design. 2. While we heavily motivate both temporal and spatial adaptivity, the subsequent breakdown of the three engines (Score, Distributed, Quantization) appears to serve only the spatial dimension, leaving the reader wondering exactly which engine executes the temporal "simulated annealing" logic.]}

\begin{itemize}
\item  We systematically characterize the spatio-temporal heterogeneity of Adam optimizer states during training, and provide both calculable metrics and empirical evidence, thereby establishing a quantitative basis for quantization strategy. 

\item  We propose \STREAM, an efficient spatio-temporal-aware framework that performs training-stage-aware score modulation and real-time identification of critical layers, enabling scalable and memory-efficient optimization in large-scale training with minimal online decision overhead.

\item  We validate \STREAM on benchmark models with up to tens of billions of parameters. Compared with industry-standard baselines such as \textit{bitsandbytes}, our method reduces optimizer memory usage by approximately 84.4\% with an average bit-width of only 5.1 bits, while maintaining comparable or superior convergence stability.
\end{itemize}

% Our code is anonymously available at XXX.

\section{Related Work}

%Low-bit quantization has become a key technique for alleviating the memory bottleneck in ultra-large-scale model training. First, 8-bit Adam \cite{dettmers20218} demonstrated the feasibility of compressing optimizer states without noticeable accuracy loss by introducing block-wise quantization. Then, to address the accuracy degradation of general quantization methods under specific operators, Jetfire \cite{xi2024jetfire} optimized the INT8 dataflow for Transformer architectures. As hardware-native support continued to evolve, the FP8 formats (E4M3/E5M2) \cite{micikevicius2022fp8} and the integrated Transformer Engine library \cite{NVIDIA_TransformerEngine_UserGuide} gradually developed from early theoretical proposals into an industrial standard for trillion-parameter model pretraining. Furthermore, to push memory reduction further, the OCP microscaling data format standard w\cite{OCP_MX_2024} introduced a finer-grained scaling mechanism. In addition, recent studies have explored sparse quantization representations such as SpQR \cite{dettmers2023spqr} in order to maintain training stability under even lower bit widths.

\textbf{Fixed-Bit Strategies}. Low-bit quantization has become a key technique for alleviating the memory bottleneck in ultra-large-scale model training. Early work such as 8-bit Adam\cite{dettmers20218} demonstrated the feasibility of compressing optimizer states with negligible accuracy loss via block-wise quantization. Subsequent efforts, including Jetfire \cite{xi2024jetfire}, further improved low-bit training by optimizing INT8 dataflow for Transformer architectures. With the evolution of hardware-native support, FP8 formats (E4M3/E5M2)\cite{micikevicius2022fp8} and the Transformer Engine library\cite{NVIDIA_TransformerEngine_UserGuide} have gradually become an industrial standard for trillion-parameter pretraining. In parallel, the OCP microscaling data format standard\cite{OCP_MX_2024} introduced a finer-grained scaling mechanism to further reduce memory usage.

However, most of these methods still follow a fixed-bit strategy. In essence, this design tends to sacrifice flexibility in exchange for deterministic operator execution efficiency. As a result, it fails to capture the complex spatiotemporal non-stationarity in gradient evolution. This limitation often leads to significant precision redundancy in the later stage of training. Moreover, such methods cannot easily achieve sub-bit-level deep compression on existing hardware platforms without native FP8 support.

% \vspace{-10pt}

\noindent\textbf{Mixed-Bit Strategies}. Mixed-precision quantization exploits structural redundancy by assigning different bit widths to different layers. For example, AnyPrecision \cite{park2024any} investigated the sensitivity differences of optimizer states under different bit widths. However, its bit-allocation strategy still mainly relies on manually designed heuristic rules. In the area of automated search, earlier studies such as HAQ \cite{wang2019haq} used reinforcement learning to search for bit widths, while the HAWQ series \cite{dong2019hawq,yao2021hawq} further introduced Hessian trace analysis to guide quantization. Later, methods such as SEAM \cite{tang2023seam}, BSQ \cite{yang2021bsq}, and ZeroQuant \cite{yao2022zeroquant} for large-scale models further improved the automation of mixed-precision allocation.

Although these search-based methods adopt mixed-precision quantization, the search space still grows rapidly with increasing model depth.
In general, the complexity can be expressed as \( \text{bits}^{\text{Layers}} \). Moreover, offline search algorithms usually make decisions based on static snapshots from the early stage of training. Therefore, they cannot effectively capture the dynamic evolution of gradients, which change from strong fluctuations in the early stage to high sparsity in the later stage. As a result, these methods ignore the continuous shift of spatiotemporal sensitivity and fail to maintain the optimal compression ratio throughout the entire training process.

\noindent\textbf{Architectural Optimization Strategies}. Beyond bit-width allocation, another line of research reduces memory overhead by reformulating the mathematical structure of the optimizer itself. Representative methods include Lion \cite{chen2023symbolic}, which removes the second-order moment in Adam-style optimizers by keeping only momentum and using sign-based updates; GaLore \cite{zhao2024galore} and its quantized version Q-GaLore \cite{zhang2024q}, which constrain the optimization process to a low-dimensional subspace through low-rank gradient projection; and Adam-mini \cite{zhang2024adam}, which exploits the approximately block-diagonal Hessian structure in Transformers to compress second-order moments or learning-rate scales from the parameter level to the block level, thereby significantly reducing memory usage. In addition, Sophia \cite{liu2023sophia}, A-LOMO \cite{lv2024full}, and MeZO \cite{malladi2023fine} explore optimizer-state compression from the perspectives of second-order approximation, training-process fusion, and zeroth-order optimization, respectively. Meanwhile, in the broader low-bit training ecosystem, methods such as QLoRA \cite{dettmers2023qlora}, DoRA \cite{liu2024dora}, BitNet \cite{wang2023bitnet,ma2024era}, OneBit \cite{xu2024onebit}, and AWQ \cite{lin2024awq} continue to push training and fine-tuning toward lower-bit model representations.

%Beyond bit-width allocation, another line of work reduces training memory by reformulating the optimizer itself. Representative approaches include removing second-order moments, as in Lion \cite{chen2023symbolic}; constraining optimization to low-rank subspaces, as in GaLore and Q-GaLore \cite{zhao2024galore,zhang2024q}; and compressing optimizer statistics from the parameter level to the block level, as in Adam-mini \cite{zhang2024adam}. Related methods further reduce optimizer overhead through second-order approximation, fused training, or zeroth-order optimization \cite{liu2023sophia,lv2024full,malladi2023fine}. More broadly, low-bit training and fine-tuning methods also push model representations toward lower precision \cite{dettmers2023qlora,liu2024dora,wang2023bitnet,ma2024era,xu2024onebit,lin2024awq}. 

However, these methods generally rely on predefined structural or heuristic rules, which limits their flexibility and fine-grained adaptability throughout training. As training dynamics evolve, they cannot promptly adjust to changing precision requirements.

%However, most of these methods rely on predefined structures for compression, and thus still struggle to balance memory efficiency with fine-grained adaptivity. In particular, Lion \cite{chen2023symbolic} weakens second-order statistical modeling, while Adam-mini sacrifices parameter-level heterogeneity. As a result, these methods still lack sufficient flexibility to handle the changing quantization demands throughout the training process.

\section{Methods}

\subsection{Problem Formulation}

We formulate memory optimization in large-scale training as a spatio-temporally constrained discrete precision allocation problem.
Consider a model $W=\{w_l\}_{l=1}^{L}$ with $L$ layers. Under adaptive optimizers such as AdamW, each layer $l$ at step $t$ maintains two momentum states: the first moment $m_{l,t}$ and the second moment $v_{l,t}$. In standard training, both states are stored in full precision (32-bit) in training, which incurs a substantial memory overhead, as shown in Equation~\eqref{eq-1}
\begin{equation}\label{eq-1}
M_{\text{t}}^{full} = \sum_{l=1}^{L} 2 \cdot N_l \cdot B_{\text{full}},
\end{equation}
where $M_t^{\mathrm{full}}$ denotes the full-precision memory overhead at setp $t$, $N_l$ is the number of parameters in layer $l$, and $B_{\text{full}}=32$ is the full precision bit-width.

To reduce this overhead, \STREAM stores optimizer states in mixed precision and selects bit-widths dynamically throughout training. Let $\mathcal{B}=\{4,8,16,32\}$ denote the set of candidate bit-widths. \STREAM aims to learn an adaptive mapping function $\mathcal{F}$ that, at each training step $t$, assigns a layer-wise bit-width $b_{l,t}\in\mathcal{B}$ based on current gradient statistics, thereby minimizing the memory footprint of optimizer states while maintaining stable convergence. Let $M_t$ denote the memory overhead at training step $t$ under adaptive bit-width allocation over $\mathcal{B}$. The problem can then be formulated as follows.
\begin{equation}
\min_{\mathcal{F}} \quad M_t = \sum_{l=1}^{L} 2 \cdot N_l \cdot b_{l,t}
\end{equation}
subject to
\begin{equation}
|\mathcal{L}(W, B_{\text{full}}) - \mathcal{L}(W, \mathcal{F})| < \epsilon,
\end{equation}
where $\mathcal{L}(W,\mathcal{F})$ denotes the loss associated with the precision allocation strategy specified by the mapping function $\mathcal{F}$.

% This formulation captures the key objective of STREAM: \textit{reducing VRAM usage by adapting precision across layers and time steps without degrading convergence}.

\subsection{The Overview of \STREAM}
\begin{figure}[t!] % 星号 * 代表横跨双栏，[t] 代表放在页顶
  \centering
  % width=\textwidth 确保图片撑满左右两栏的总宽度
  \includegraphics[scale=0.35]{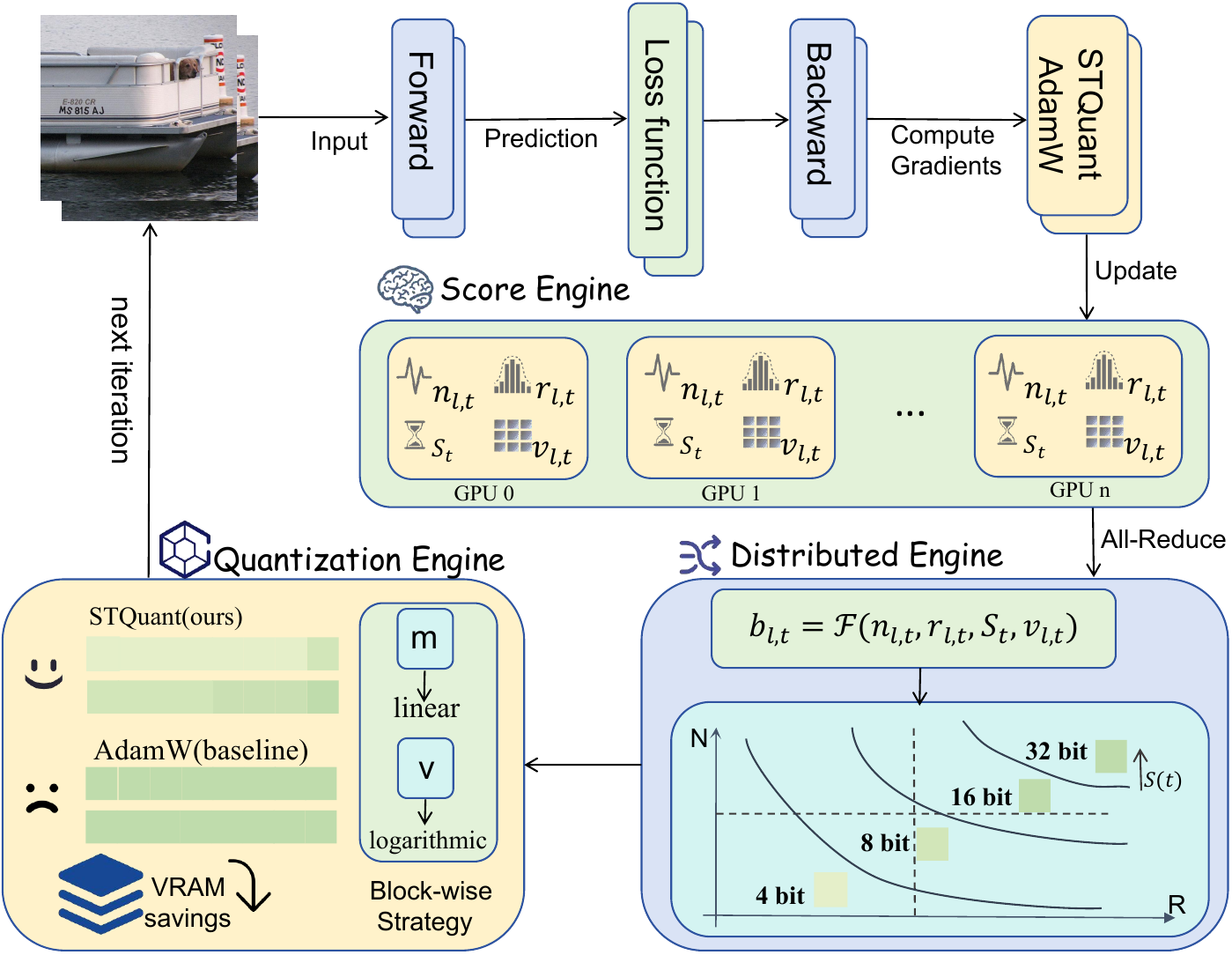} 
  \caption{Overview of the \STREAM framework: The system consists of three collaborative engines: (1) the \textbf{Score Engine} extracts spatio-temporal gradient features across GPUs; (2) the \textbf{Distributed Engine} synchronizes global statistics to determine optimal bit-widths; and (3) the \textbf{Quantization Engine} executes dual-mode block-wise compression (for $m$ and $v$).}
  \label{fig:method}
\end{figure}

Figure~\ref{fig:method} illustrates the overall workflow of \STREAM, which primarily comprises three coordinated engines: (1) \textbf{Score Engine} computes multi-dimensional gradient statistics, including $n_{l,t}, r_{l,t}, s_t,$ and $v_{l,t}$, in parallel across all GPU nodes; (2) \textbf{Distributed Engine} is responsible for synchronizing gradient statistics and utilize the mapping function $\mathcal{F}$ to dynamically determine the optimal bit-width $b_{l,t} \in \mathcal{B}$ for each layer, thereby ensuring consistency across the distributed environment; (3) \textbf{Quantization Engine} performs block-wise quantization of the optimizer states based according to the assigned bit-widths. Specifically, \STREAM applies linear mapping to the first moment, while using logarithmic quantization for the second moment to accommodate its larger numerical range.
% As shown in the bottom-left of Figure~\ref{fig:method}, STREAM significantly reduces VRAM overhead compared to the standard AdamW baseline.

Notably, \STREAM introduces only $\mathcal{O}(1)$ auxiliary memory overhead, which means the storage overhead does not grow linearly with the number of parameters $N$ or layers $L$. Moreover, despite the complexity of the search space, the decision-making overhead remains negligible compared with the total training time, even for trillion-parameter models.

% In contrast, traditional mixed-precision quantization schemes often attempt to find a static optimal bit-combination, resulting in a search space complexity of $\mathcal{O}(|\mathcal{B}|^L)$. STREAM effectively transforms this complex combinatorial optimization problem into a low-overhead dynamic decision process, significantly reducing the implementation complexity of dynamic quantization.

 % 别忘了结尾也要加星号

\subsection{Score Engine}\label{score-engine}
\subsubsection{Bi-factor Sensitivity Proxy}

% \textbf{Feature Factor Definition:} 
To construct an efficient precision mapping function, \STREAM avoids the massive computational overhead associated with the Hessian matrix. Instead, it leverages first-order gradient statistics as a lightweight proxy for second-order sensitivity. For each layer $l$, we define two complementary feature descriptors at the current time step $t$: the Intensity Factor ($n$) and the Variation Factor ($r$).

$\bullet$ Intensity Factor: We define $n_{l,t}$ in Equation~\eqref{eq-2} as the Root Mean Square (RMS) of the gradient elements, characterizing the overall gradient magnitude and the sensitivity scale of the gradient by layer:
    \begin{equation}\label{eq-2}
        n_{l,t} = \sqrt{\frac{1}{N_l} \sum_{i=1}^{N_l} (g_i)^2}
    \end{equation}
    
$\bullet$ Variation Factor:  We define $r_{l,t}$ in Equation~\eqref{eq-3} as the Coefficient of Variation (CV) of the gradient magnitudes, measuring the dispersion and heterogeneity of the gradient distribution:
    \begin{equation}\label{eq-3}
        r_{l,t} = \frac{\sigma(|g_{l,t}|)}{\mu(|g_{l,t}|) + \epsilon}
    \end{equation}

To eliminate instantaneous random fluctuations during training and capture long-term statistical patterns, we introduce the Exponential Moving Average (EMA) to maintain historically smoothed estimate statistics $N_{\text{ema}}$ and $R_{\text{ema}}$:
\begin{align}\label{eq-4}
    N_{\text{ema}}^{(t)} &= \alpha \cdot \text{Mean}(\{n_{l,t}\}_{l=1}^L) + (1-\alpha)N_{\text{ema}}^{(t-1)} \\
    R_{\text{ema}}^{(t)} &= \alpha \cdot \text{Mean}(\{r_{l,t}\}_{l=1}^L) + (1-\alpha)R_{\text{ema}}^{(t-1)}
\end{align}

\textbf{Theoretical Analysis:} \STREAM adopts $n$ and $r$ as the decision basis. The core idea lies in using the Fisher Information Matrix (FIM) theory to perform a lightweight approximation of Hessian-trace-based sensitivity metrics (e.g., HAWQ-V2 \cite{dong2020hawq}). Specifically,  under standard FIM theory, the expected Hessian can be approximated by the expected outer product of gradients during training, i.e., $\mathbb{E}[\mathbf{H}] \approx \mathbb{E}[\mathbf{gg}^T]$. Therefore, we take the trace of both sides as Equation~\eqref{eq-5}, indicating that the second-order sensitivity of a layer is proportional to the square of the Frobenius norm of its gradient:
\begin{equation}\label{eq-5}
    \mathbb{E}[\text{Tr}(\mathbf{H})] \propto \mathbb{E}[\|g\|_F^2]
\end{equation}

\begin{figure}[t!] 
    \centering
    \hspace*{2.5em} 
    \includegraphics[scale=0.45]{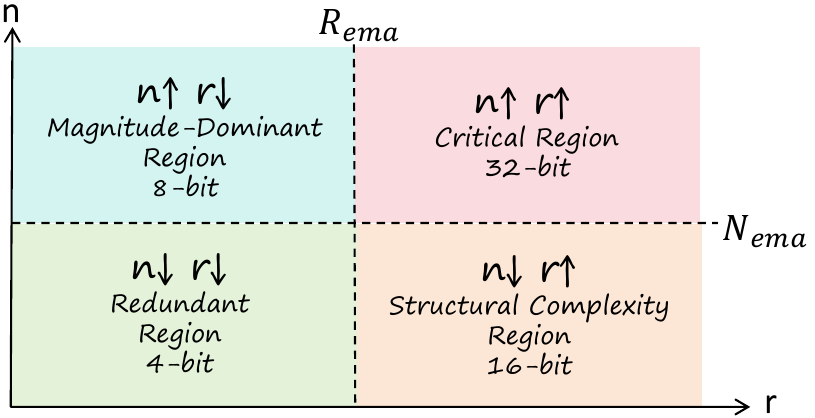} % 调整宽度比例，0.8 比较合适
    \caption{The $n$-$r$ decision quadrants for bit-width allocation. The decision space is partitioned into four zones based on global EMA statistics ($N_{\text{ema}}$ and $R_{\text{ema}}$): (1) \textit{Critical Zone} (top-right), (2) \textit{Magnitude-Dominant Zone} (top-left), (3) \textit{Structural Complexity Zone} (bottom-right), and (4) \textit{Redundant Zone} (bottom-left).}
    \label{fig:nr_quadrants}
\end{figure}

FurtherMore, for a fully connected layer $y = Wx$, the gradient is define as $\nabla_W \mathcal{L} = \frac{\partial \mathcal{L}}{\partial y} \cdot x^T$, and the Frobenius norm satisfies:
\begin{equation}\label{eq-6}
    \|\nabla_W \mathcal{L}\|_F^2 = \left\| \frac{\partial \mathcal{L}}{\partial y} \right\|_2^2 \cdot \|x\|_2^2
\end{equation}
Equation~\eqref{eq-6} implies that the weight gradient norm captures the joint scale evolution of activation magnitudes and error signals. The intensity factor $n$, as a normalized expression of the gradient norm, characterizes the overall scale of the Hessian trace. However, $n$ alone cannot reflect the distribution characteristics of the \textbf{Hessian Spectrum}. Thus, we introduce the variation factor $r$ to further depict intra-layer heterogeneity in parameter sensitivity. As illustrated in Figure~\ref{fig:nr_quadrants}, we map each layer into the $n$-$r$ decision quadrants to determine the corresponding bit-width allocation logic. Consequently, the combination of $n$ and $r$ enables a high-fidelity approximation consistent with HAWQ-V2~\cite{dong2020hawq} with minimal computational cost.

% Based on the bivariate space formed by $n$ and $r$, as illustrated in Figure~\ref{fig:nr_quadrants}, we project each layer into the $n$-$r$ decision quadrants to determine the corresponding bit-width allocation logic.

\subsubsection{Temporal Annealing Factor}

While the bi-factor descriptors effectively capture spatial sensitivity, they remain inherently instantaneous and local observations. To improve the robustness of our decisions, we further incorporate prior knowledge of training dynamics into the \STREAM framework. As illustrated in Figure~\ref{fig:motivation}, deep learning training is a non-stationary process that evolves from chaotic exploration toward local convergence. In the early stages of pre-training, parameters are randomly initialized, leading to unstable gradient directions and severe amplitude fluctuations. Applying low-bit quantization at this stage may cause quantization noise to be amplified layer by layer through non-linear mappings, potentially interfering with the convergence trajectory or even causing numerical divergence. To mitigate this, we introduce the temporal annealing factor ($S_t$) as a numerical stability buffer during the initial phase of training, and parameterize its schedule using the hyperbolic secant function, defined as Equation~\eqref{eq-7}:
% We adopt the hyperbolic secant function as the basis for the scheduling curve, defined as:
\begin{equation}\label{eq-7}
    S_t = 1 + \text{sech}\left(\frac{t}{\tau}\right)
\end{equation}
where $t$ is the current training step and $\tau$ is an adaptive time constant that controls the decay rate of $S_t$, i.e., the length of the high-precision protection window. We set $\tau$ adaptively to account for variations across architectures and training settings: deeper models with $L$ layers typically accumulate errors more strongly and require more iterations for gradients to stabilize, whereas larger batch sizes reduce the variance of gradient estimates under the square-root scaling rule~\cite{hoffer2017train}, changing the effective information gain per step. Consequently, Equation~\eqref{eq-7} exhibits several desirable properties that match the requirements of precision scheduling:

$\bullet$ Initial Inertia: Around $t=0$, $S'(0)=0$, the derivative $S'(0) = 0$, which enables a smooth warm-up of the bit-width policy during the startup phase and prevents precision mutations from undermining cold-start stability.

$\bullet$ High-order Continuity: $S_t$ is twice-differentiable, guaranteeing that bit-width switching boundaries evolve continuously and smoothly over time.
 
$\bullet$ Asymptotic Decay: As $t$ increases, $\text{sech}(t/\tau)$ decays exponentially toward 0, allowing $s_t$ to naturally transition from the protection mode back to the feature-driven mode.

% Considering that the demand for precision protection varies across model architectures and training tasks, we define $\tau$ as an adaptive time constant. On one hand, deeper models ($L$) exhibit more pronounced error accumulation effects, requiring more iterations for gradients to stabilize. On the other hand, according to the square root scaling rule \cite{hoffer2017train}, larger batch sizes reduce the variance of gradient estimation and alter the information gain per step. Therefore, by coupling the model depth ($L$) with the batch size, $\tau$ ensures that STREAM adaptively defines the optimal high-precision protection period based on the specific task scale and architecture.
\subsubsection{Hierarchical Feature Factor}

From a spatial perspective, to characterize the varying sensitivity of different layers to quantization errors, we introduce the hierarchical sensitivity factor as an evaluation metric. Let $g_{i,t}$ denote the gradient of parameter $i$ at step $t$. We quantify the the $l$-th layer gradient scale at the current update step, $v_{l,t}$, as follows:
\begin{equation}\label{eq-8}
    v_{l,t} = \text{Mean}(\{g_{i,t}^2\}_{i \in \text{Layer } l})
\end{equation}
 Equation~\eqref{eq-8} intuitively reflects the overall gradient strength and parameter activity within the layer. However, functional modules in deep neural networks (e.g., Transformers), such as self-attention and MLP, exhibit inherent magnitude heterogeneity in their gradient distributions. Consequently, relying solely on raw magnitude statistics fails to achieve equitable precision allocation across the entire model.

To address this, we maintain a dynamic, historically smoothed estimate, $V_{\text{global\_ema}}$, which effectively eliminates inter-layer magnitude discrepancies. Specifically, we first aggregate the layer-wise magnitude statistics across all $L$ layers and apply the Exponential Moving Average (EMA):
\begin{equation}
    V_{\text{global\_ema}}^{(t)} = \alpha \cdot \frac{1}{L} \sum_{l=1}^{L} v_{l,t} + (1 - \alpha) V_{\text{global\_ema}}^{(t-1)}
\end{equation}
By introducing a historically smoothed, we further define the ratio $v_{l,t} / V_{\text{global\_ema}}$ as an evaluation of the relative importance of a layer compared to the entire model. As a result, \STREAM can ensure that limited memory resources are optimally scheduled across layers, prioritizing those critical layers whose magnitude fluctuations significantly exceed the global average.

\subsection{Distributed Engine}
As defined in Section ~\ref{score-engine}, the four statistical features $n_{l,t}$, $r_{l,t}$, $s_{t}$, and $v_{l,t}$ jointly characterize the quantization sensitivity of each layer. Since the final bit-width allocation requires a unified ranking criterion, we aggregate these features into a single scalar score. Motivated by \textit{Rate-Distortion} theory\cite{shannon1959coding, cover1999elements},we follow a simple principle. The additional bit-width assigned to each layer depends on its relative sensitivity with respect to a global reference. In particular, rate-distortion analysis indicates that the required increase in bit precision is proportional to the logarithm of the signal variance or sensitivity. Therefore, if a layer is $k$ times more sensitive than the global average, its bit-allocation score should increase by approximately $\log_2(k)$.
Following this intuition, we first normalize layer-wise statistics by their corresponding global running estimates, so that each term measures a relative amplification factor rather than an absolute magnitude. We then apply the logarithm to convert multiplicative deviations into additive contributions, yielding the unified scoring function:
\begin{equation}
\begin{split}
    \text{score}_{l,t} = \Phi & + \log_2 \frac{r_{l,t}}{R_{\text{ema}}^{(t)}} + \log_2 \frac{n_{l,t}}{N_{\text{ema}}^{(t)}} \\
    & + \log_2 S(t) + \log_2 \frac{v_{l,t}}{V_{\text{global\_ema}}^{(t)}}
\end{split}
\end{equation}

This scoring function has two advantages. First, each term has a clear interpretation: a larger ratio indicates that the current layer is more sensitive than the global baseline in that aspect. Second, the logarithmic form makes the contributions additive, which enables a simple and interpretable fusion of multiple heterogeneous statistics. As a result, layers with consistently larger relative sensitivity obtain higher scores and are assigned higher bit-widths.

Finally, the continuous score is projected into a discrete bit-width space through a step-wise mapping function $\mathrm{Map}(\cdot)$:
\begin{equation}
    b_{l,t} = \text{Map}(\text{score}_{l,t}) \in \{4, 8, 16, 32\}
\end{equation}

The hyperparameters within the mapping function $\text{Map}(\cdot)$ are determined through sensitivity distribution profiling across representative multimodal models. The base bias $\Phi = 7.2$ is utilized to anchor the 8-bit reference precision. The thresholds $\{6.8, 12, 24\}$ are established based on the quantile statistics of gradients during the training process. This design enables differentiated resource scheduling: non-critical layers at the lower end of the sensitivity distribution are compressed to 4-bit to maximize VRAM savings, whereas the critical layers---characterized by violent gradient dynamics and high sensitivity to long-term cumulative errors---are assigned higher bit-widths to ensure training stability.
\begin{algorithm}[t]
\caption{Spatio-Temporal Adaptive Bit-Width Allocation}
\label{alg:mdq_adamw}
\begin{algorithmic}[1]
\REQUIRE Learning rate $\eta$, weight decay $\lambda$, coefficients $\beta_1, \beta_2$, EMA factor $\alpha$, decay period $\tau$, update frequency $U$, block size $B$, small constant $\epsilon$.
\STATE Initialize step $t \leftarrow 0$, global EMA stats $\bar{n}, \bar{r}, \bar{v}_{global} \leftarrow 0$.
\STATE Initialize layer states $m_0, v_{0,sq} \leftarrow 0$, initial bit-width $b_l \leftarrow 16$ for each layer $l$.
\WHILE{training not converged}
    \STATE $t \leftarrow t + 1$
    % \STATE Compute stochastic gradient 
    $g_t = \nabla f_t(\theta_{t-1})$
    \IF{($t \bmod U = 0$) \OR ($t < 5$)}
        \FOR{each layer $l = 1, \dots, L$}
            % \STATE Compute local stats:
            \STATE $n_l = \sqrt{\mathbb{E}[g_{l,t}^2]}$, \ $r_l = \sigma(g_{l,t}) / (\mathbb{E}[|g_{l,t}|] + \epsilon)$, \ $v_l = \mathbb{E}[g_{l,t}^2]$
        \ENDFOR
        % \STATE Aggregate and update global EMA (cross-device):
        \STATE $\bar{n} \leftarrow \alpha \cdot \text{avg}(n_l) + (1-\alpha)\bar{n}$
        \STATE $\bar{r} \leftarrow \alpha \cdot \text{avg}(r_l) + (1-\alpha)\bar{r}$
        \STATE $\bar{v}_{global} \leftarrow \alpha \cdot \text{avg}(v_l) + (1-\alpha)\bar{v}_{global}$
        \FOR{each layer $l$}
            % \STATE Compute decay factor: 
            \STATE $s_t = 1 + \text{sech}(t/\tau)$
            % \STATE Calculate allocation score:
            \STATE $Score_l = 7.2 + \log_2 \frac{r_l}{\bar{r} + \epsilon} + \log_2 \frac{n_l}{\bar{n} + \epsilon} + \log_2 s_t + \log_2 \frac{v_l}{\bar{v}_{global} + \epsilon}$
            % \STATE Determine bit-width $b_l \in \{4, 8, 16, 32\}$ via threshold mapping:
            \STATE $b_l = \text{Map}(Score_l)$ (Thresholds: 6.8, 12, 24)
        \ENDFOR
    \ENDIF
    \FOR{each layer $l$}
        % \STATE Update momentum estimates:
        \STATE $m_t = \beta_1 m_{t-1} + (1-\beta_1)g_t$
        \STATE $v_{t,sq} = \beta_2 v_{t-1,sq} + (1-\beta_2)g_t^2$
        % \STATE \textbf{Robust Quantization:}
        \STATE $\hat{m}_t = \text{Quantize}(m_t, b_l, \text{mode='linear'}, \text{block}=B)$
        \STATE $\hat{v}_{t,sq} = \text{Quantize}(v_{t,sq}, b_l, \text{mode='log'}, \text{block}=B)$
        % \STATE Bias correction: 
        \STATE $\tilde{m}_t = \hat{m}_t / (1-\beta_1^t), \ \tilde{v}_{t,sq} = \hat{v}_{t,sq} / (1-\beta_2^t)$
        % \STATE Weight update: 
        \STATE $\theta_t = \theta_{t-1} \cdot (1 - \eta \lambda) - \eta \cdot \frac{\tilde{m}_t}{\sqrt{\tilde{v}_{t,sq}} + \epsilon}$
    \ENDFOR
\ENDWHILE
\end{algorithmic}
\end{algorithm}

\subsection{Quantization Engine}
After determining the specific bit-width $b_{l,t}$ for each layer, we adopt a block-wise quantization strategy that partitions parameter tensors into several contiguous sub-blocks. By localizing the scaling factors, we can effectively suppress quantization noise and prevent numerical outliers from distorting the global quantization scale.

To match the different statistical characteristics of optimizer states, we use a dual-mode quantization scheme:

$\bullet$  \textbf{First Moment ($m$):} We apply linear symmetric quantization. Since the first moment carries the directional information of gradient descent, linear mapping preserves the integrity of the optimization trajectory.

$\bullet$ \textbf{Second Moment ($v$):} In contrast, we apply logarithmic quantization to the second moment. Because squared gradients typically exhibit a vast dynamic range and a heavy-tailed distribution, mapping them into the logarithmic domain effectively compresses the numerical span. This method allows us to capture fine changes in small values even at very low bit-widths.

Through a spatio-temporally adaptive mechanism, \STREAM enables an intelligent quantization strategy. Algorithm~\ref{alg:mdq_adamw} summarizes how \STREAM works during training. First, we compute the gradient and periodically collect layer-wise statistics(Line 4-7). These statistics are aggregated across GPUs and smoothed by EMA to obtain the historically smoothed estimate. Then, we use them to assign a bit-width to each layer(Line 9-11). The allocation score combines spatial differences across layers and temporal changes over training. The resulting bit-width is selected from $\{4,8,16,32\}$ (Line 13-15).
After that, we update the first and second-order states of the optimizer. The first moment $m$ is quantized with a linear scheme, while the second moment $v$ is quantized with a logarithmic scheme(Line 19-22). Finally, we use the quantized states to update bias and model parameters (Line 23-24).

%\vspace{-10pt}

\section{Experiments}

\subsection{Baselines and settings}
\textbf{Baselines.} We evaluate STAF-Q against 32-bit AdamW~\cite{loshchilov2017decoupled} and bitsandbytes 8-bit AdamW~\cite{dettmers20218} across pre-training, fine-tuning, and ablation experiments in language, vision, and image-text retrieval. The former serves as a full-precision baseline and the latter as a practical low-bit baseline. Benchmarks, models, and evaluation metrics are summarized in Table~\ref{tab:settings_overview}.

%We evaluate \STREAM AdamW against 32-bit AdamW\cite{loshchilov2017decoupled} and 8-bit AdamW\cite{dettmers20218} across a diverse set of pre-training, fine-tuning, and ablation experiments spanning language, vision, and image-text retrieval. The corresponding benchmarks, models, and evaluation metrics are summarized in Table~\ref{tab:settings_overview}.

% Unless otherwise noted, all experiments are conducted on NVIDIA A800 (80GB) GPUs using FP16 training. Pre-training is performed on 4 GPUs, whereas fine-tuning and ablation are conducted on a single GPU. For fair comparison, all training settings are kept identical across optimizers, including random seeds and hyperparameters, with optimizer state representation being the only difference.

\noindent\textbf{Settings.} All experiments are conducted on NVIDIA A800 (80GB) GPUs with FP16 training. Pre-training uses four GPUs, while fine-tuning and ablation studies use a single GPU. To ensure a fair comparison, we keep all training settings identical across optimizers, including random seeds and hyperparameters, and vary only the optimizer-state representation.

\begin{table}[t!]
\centering
\caption{Overview of experimental settings.}
\label{tab:settings_overview}
\small
\setlength{\tabcolsep}{4pt}
\renewcommand{\arraystretch}{1.15}
\resizebox{\columnwidth}{!}{
\begin{tabular}{c c c c c}
\toprule
\rowcolor{gray!18}
\textbf{Stage} & \textbf{Task} & \textbf{Data} & \textbf{Model} & \textbf{Metric} \\
\midrule
\rowcolors{2}{gray!8}{white}
Pre-train & LM     & OpenWebText  & GPT2-XL             & Loss \\
Pre-train & VRecon & COCO 2017    & ViT-Base            & Loss / Top-1 \\
Fine-tune & NLU    & MNLI         & RoBERTa-Large       & Accuracy \\
Fine-tune & Vision & COCO 2017    & ViT-Base            & mAP \\
Fine-tune & LM     & Wikitext-103 & GPT2-Medium         & PPL \\
Fine-tune & ITR    & COCO 2017    & ViT-B/32            & Recall@1 \\
Ablation  & LM     & Wikitext-103 & 12-layer GPT Trans. & PPL \\
\bottomrule
\end{tabular}
}

%\vspace{1mm}
\parbox{\columnwidth}{\footnotesize\textit{Note:} LM = Language Modeling; VRecon = Visual Reconstruction; NLU = Natural Language Understanding; Vision = Visual Recognition; ITR = Image-Text Retrieval.}
\end{table}

\subsection{Pre-training Analysis}
In the pre-training stage, we focus on four key aspects of \STREAM: convergence stability, cross-modal generalization, the quality of pre-trained representations, and the memory efficiency of optimizer states. To this end, we analyze its convergence behavior on both language and vision pre-training tasks. In addition, we use linear probing results, memory comparisons, and bit-width evolution plots to examine its performance. Through these evaluations, we verify whether \STREAM can maintain both training quality and resource efficiency under an extremely low average bit-width.

%\vspace{-5pt}

\subsubsection{Convergence in Language and Vision Pre-Training Tasks}
\begin{figure}[t!]
    \centering
    \includegraphics[width=0.98\columnwidth]{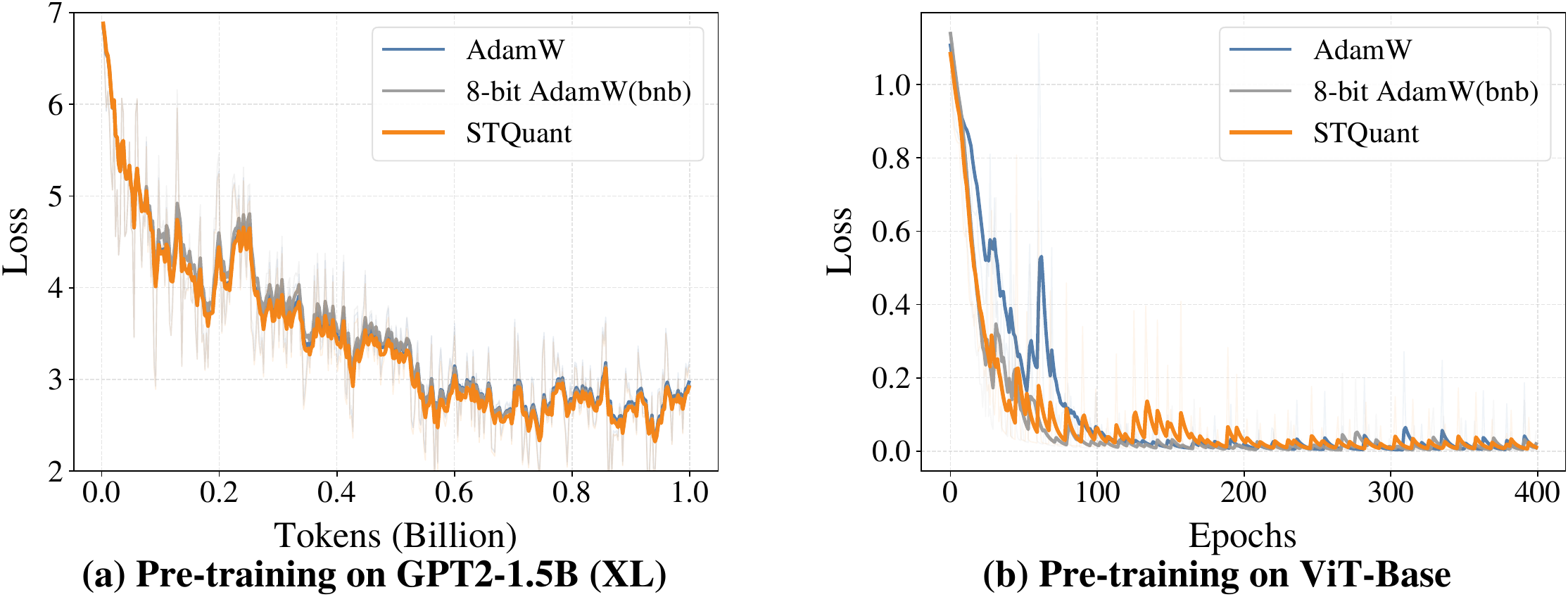}
    \caption{Pre-training on GPT2-1.5B (XL) and ViT-Base.}
    \label{fig:pretraingpt}
\end{figure}

We first study the convergence behavior of \STREAM under different pre-training tasks.
Figure~\ref{fig:pretraingpt} shows that the training loss curve of \STREAM closely matches that of full-precision AdamW during pre-training.
This result indicates that \STREAM maintains an optimization trajectory highly consistent with that of the full-precision baseline, despite significant compression of the optimizer states. 
By comparison, 8-bit AdamW shows larger loss fluctuations in the early training stage and slightly weaker overall stability. This indicates that the dynamic bit-width allocation of \STREAM is better suited to the numerical stability requirements of early pre-training, thereby reducing the optimization disturbances introduced by static low-bit quantization at critical stages.

To examine whether this advantage extends to the visual modality, we further analyze the vision pre-training task. As shown in Figure ~\ref{fig:pretraingpt}, even for the vision reconstruction task, which demands stronger optimizer stability, \STREAM remains highly consistent with full-precision AdamW in terms of convergence behavior, while using an average state bit-width of only 5.1 bits. These results show that the adaptive bit-width of \STREAM is not limited to language models. Instead, it remains effective across different modalities and training objectives.
\begin{figure}[t!]
    \centering
    % 将 0.8\textwidth 改为 1.0\columnwidth 或更小
    \includegraphics[scale=0.42]{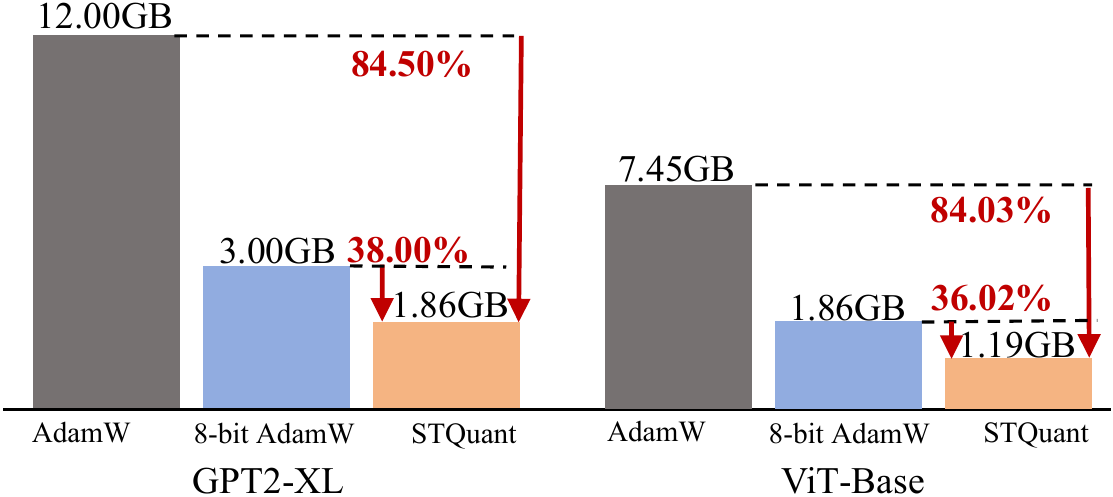} 
    \caption{Comparison of optimizer-state memory on GPT2-XL and ViT-Base. \STREAM achieves the lowest memory overhead among all compared optimizers.}
    \label{fig:optimizer}
\end{figure}

\subsubsection{Quality of Pre-trained Representations}
\begin{figure*}[t] % [t] 表示强制放在页面顶部，figure* 在 ACM 中通常只能放在顶部或单独一页
  \centering
  
  % 左侧子图
  \begin{subfigure}{0.46\textwidth}
    \centering
    \includegraphics[width=\linewidth]{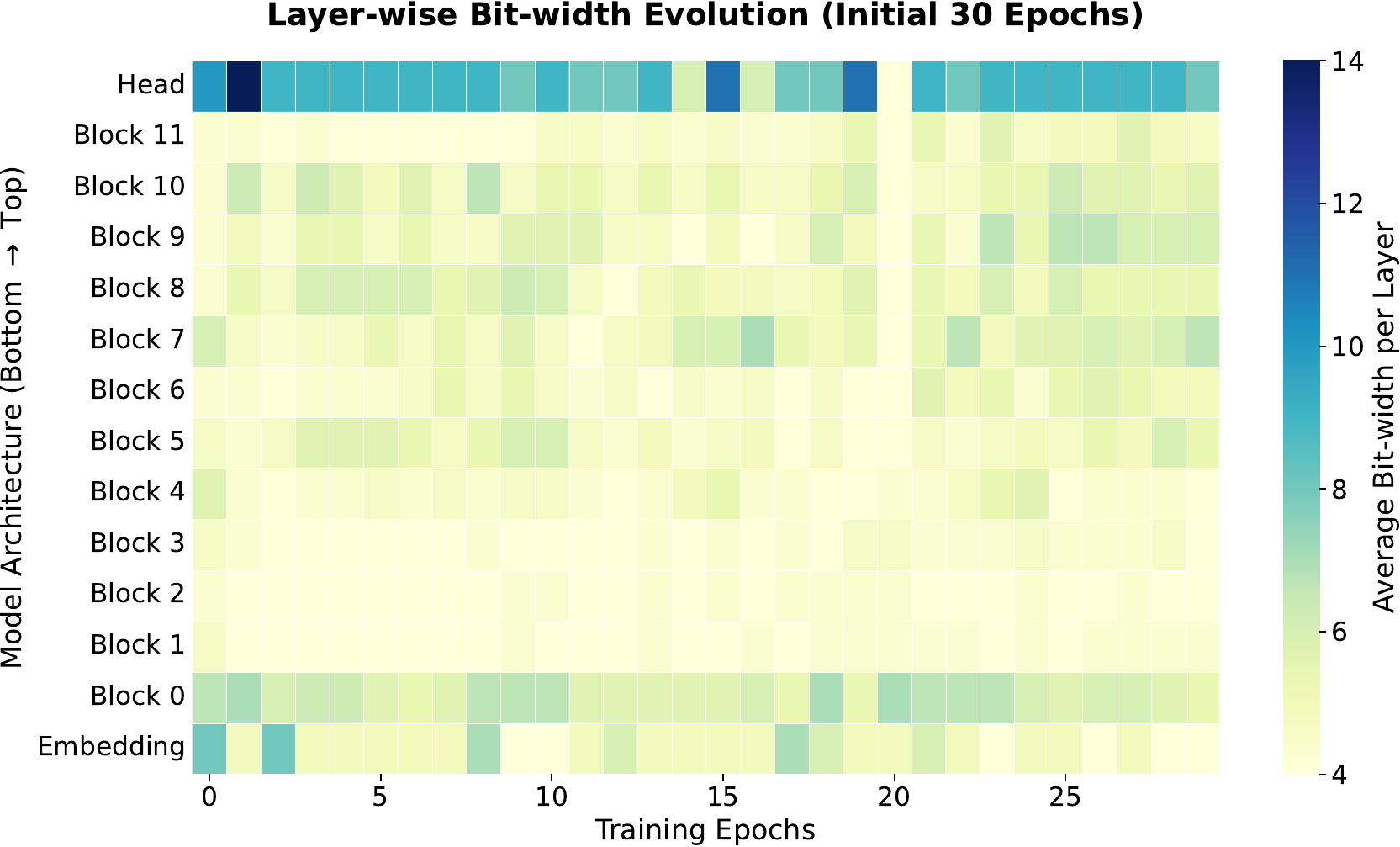}
    \caption{Macro-level layer-wise bit-width evolution.} % 子图 a 的标题
    \label{fig:layerwise}
  \end{subfigure}
  \hfill % 撑开中间的空白
  % 右侧子图
  \begin{subfigure}{0.46\textwidth}
    \centering
    \includegraphics[width=\linewidth]{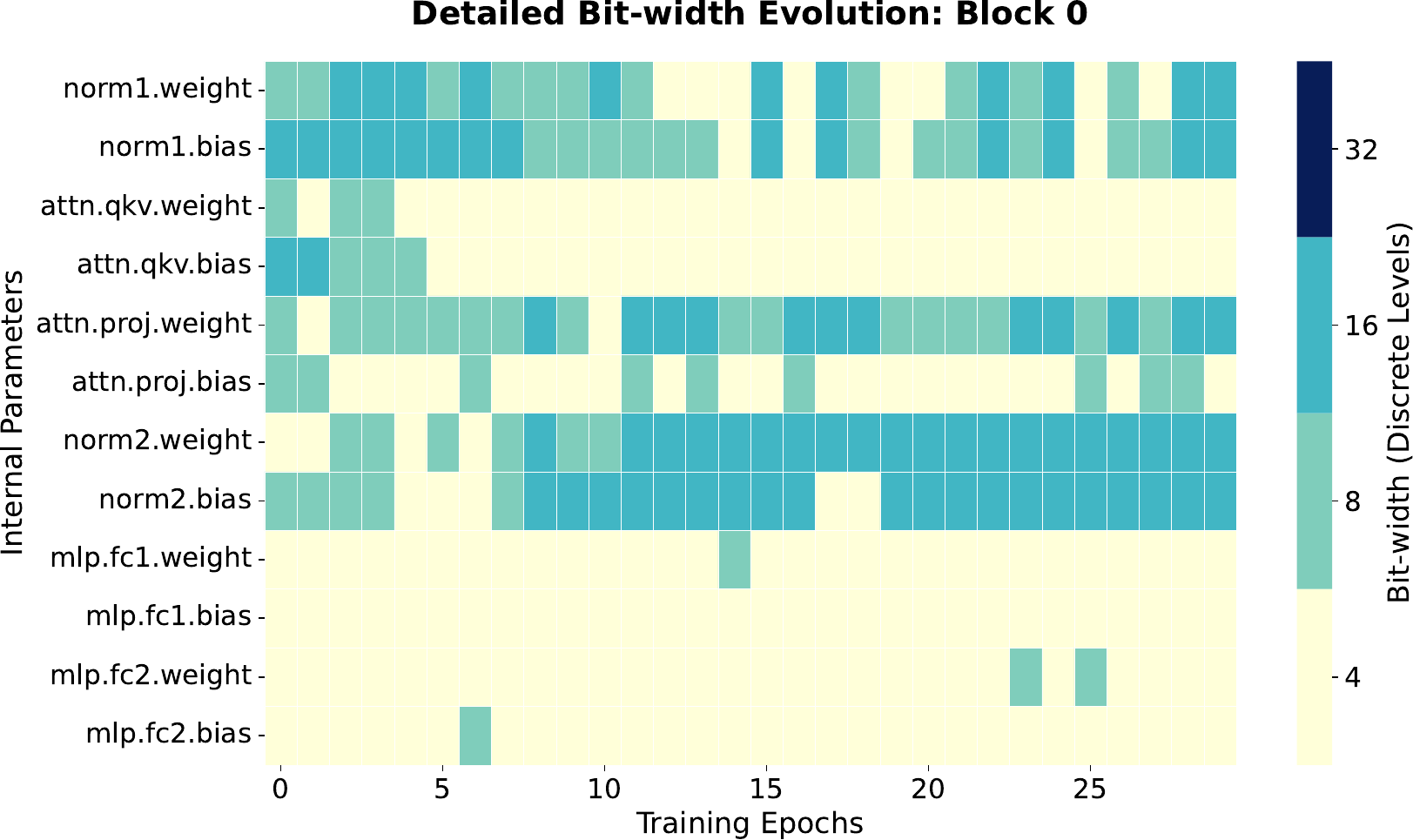}
    \caption{Micro-level bit-width evolution within a Transformer block.} % 子图 b 的标题
    \label{fig:block0}
  \end{subfigure}

  \caption{Bit-width evolution of \STREAM during pre-training. (a) shows the dynamic bit-width allocation across layers over training epochs, and (b) presents the bit-width evolution of different parameter groups within a Transformer block.} % 总标题
  \label{fig:combined_results}
\end{figure*}
Loss curves alone are not sufficient to demonstrate that a model has learned high-quality representations during pre-training. To further evaluate the representation quality after pre-training, we examine the downstream performance of the vision model using linear probing.
\begin{table}[t!]
\centering
\caption{Linear probing of ViT-Base with different optimizers.}
\label{tab:linear_probe_vit_base}
\small
\setlength{\tabcolsep}{12pt}
\begin{tabular}{cccc}
\toprule
\multicolumn{1}{c}{Optimizer} & \multicolumn{1}{c}{State} & \multicolumn{1}{c}{Avg.} & \multicolumn{1}{c}{Top-1 (\%)} \\
\midrule
32-bit AdamW & 32-bit & 32.0 & 31.44 \\
8-bit AdamW (bnb) & 8-bit & 8.0 & 31.68 \\
\rowcolor{gray!15}
\textit{STQuant} & Dynamic & 5.1 & 31.64 \\
\bottomrule
\end{tabular}
\end{table}

Table ~\ref{tab:linear_probe_vit_base} shows that the model pre-trained with \STREAM achieves a Top-1 accuracy of 31.64\%, which is on par with full-precision AdamW at 31.44\% and 8-bit AdamW at 31.68\%. This result indicates that \STREAM substantially reduces the bit-width of optimizer states without impairing the model's semantic representation ability.
In other words, \STREAM removes redundant precision in optimizer states rather than information that is essential for downstream performance.
Therefore, the benefit of \STREAM is reflected not only in its training dynamics, where the loss curve closely matches the full-precision baseline, but also in the quality of its final learned representations. This is important because it shows that \STREAM does not simply achieve an appealing training loss; it also preserves the model's representational strength.

%\vspace{-10pt}

\subsubsection{Memory Efficiency}

In addition to maintaining training quality and representation capability, \STREAM also brings significant savings in the memory overhead of optimizer states.
As shown in Figure ~\ref{fig:optimizer}, \STREAM uses much less optimizer-state memory than full-precision AdamW on both GPT2-XL and ViT-Base. Moreover, it further reduces memory usage compared with 8-bit AdamW(bnb). Specifically, on GPT2-XL, \STREAM reduces the optimizer-state memory from 12.00 GB with AdamW to 1.86 GB. On ViT-Base, it reduces the memory from 7.45 GB to 1.19 GB.  Overall, \STREAM requires only about $1/6$ of the optimizer-state memory of full-precision AdamW; Compared to 8-bit AdamW (bnb), it further reduces memory overhead by 38\% on GPT2-XL and 36.02\% on ViT-Base.

These results demonstrate that \STREAM achieves more aggressive memory compression than static 8-bit quantization while introducing almost no degradation in training performance. 
% Therefore, \STREAM is not only practically viable but also a more appealing alternative. 
In particular, in GPU memory-constrained training environments, \STREAM can significantly enhance the feasibility of large-scale pre-training.

\begin{table*}[t]
\centering
\captionsetup{justification=centering}
\caption{Performance and optimizer-state memory savings of \STREAM AdamW compared with full-precision and 8-bit AdamW(bnb) baselines across core fine-tuning tasks.}
\label{tab:stream_finetune}
\renewcommand{\arraystretch}{1.12}
\setlength{\tabcolsep}{9pt}
\resizebox{0.98\textwidth}{!}{%
\begin{tabular}{c|c c c c|c|c}

\hline
\rowcolor{gray!15}
 & \multicolumn{4}{c|}{\rule{0pt}{3.5ex}\textbf{Task Setting}} & & \\
\cline{2-5}
\rowcolor{gray!15}
% 这里增加了 [2pt] 来微调高度，使其上移
\multirow{-2}{*}[2pt]{\textbf{Optimizer}} 
& \rule{0pt}{3.5ex}\textbf{Task} 
& \textbf{Data} 
& \textbf{Model} 
& \textbf{Metric} 
& \multirow{-2}{*}[2pt]{\textbf{Value}} 
& \multirow{-2}{*}[2pt]{\makecell{\textbf{Opt.State Mem} \\ \textbf{Saved}}} \\
\hline

32-bit AdamW & GLUE & MNLI & RoBERTa-Large & Accuracy($\uparrow$) & 0.9060 & reference \\
8-bit AdamW(bnb) & GLUE & MNLI & RoBERTa-Large & Accuracy($\uparrow$) & 0.9002 & 75.0\% \\
\rowcolor{gray!12}
\STREAM & GLUE & MNLI & RoBERTa-Large & Accuracy($\uparrow$) & \textbf{0.9032} & \textbf{84.4\%} \\
\hline
32-bit AdamW & Classification & COCO 2017 & ViT-Base & mAP($\uparrow$) & 0.72669 & reference \\
8-bit AdamW(bnb) & Classification & COCO 2017 & ViT-Base & mAP($\uparrow$) & 0.73082 & 75.0\% \\
\rowcolor{gray!12}
\STREAM & Classification & COCO 2017 & ViT-Base & mAP($\uparrow$) & \textbf{0.72434} & \textbf{83.0\%} \\
\hline
32-bit AdamW & LM & wikitext-103 & GPT2-Medium & PPL($\downarrow$) & 20.02 & reference \\
8-bit AdamW(bnb) & LM & wikitext-103 & GPT2-Medium & PPL($\downarrow$) & 20.22 & 75.0\% \\
\rowcolor{gray!12}
\STREAM & LM & wikitext-103 & GPT2-Medium & PPL($\downarrow$) & \textbf{20.1} & \textbf{82.5\%} \\
\hline
32-bit AdamW & IT Retrieval & COCO 2017 & ViT-B/32 & Recall@1($\uparrow$) & 0.7240 & reference \\
8-bit AdamW(bnb) & IT Retrieval & COCO 2017 & ViT-B/32 & Recall@1($\uparrow$) & 0.7220 & 75.0\% \\
\rowcolor{gray!12}
\STREAM & IT Retrieval & COCO 2017 & ViT-B/32 & Recall@1($\uparrow$) & \textbf{0.7320} & \textbf{87.5\%} \\
\hline
\end{tabular}%
}
\end{table*}

\subsubsection{Bit-width Evolution Analysis}

To understand why \STREAM remains effective under an extremely low average bit-width, we analyze the evolution of its bit-width allocation during pre-training.

From a layer-wise view, Figure~\ref{fig:layerwise} shows that the Embedding and Head layers usually keep higher bit-widths throughout training, while the intermediate Transformer blocks are compressed more aggressively. This reveals a clear pattern: the layers at both ends are more sensitive, whereas the middle layers are more redundant. Hence, different layers have different precision requirements, and \STREAM can automatically reserve more precision for the critical ones.
From a temporal view, the bit-width of each layer does not decrease monotonically. Instead, it changes dynamically, with multiple rebounds and reallocations during training. This means that \STREAM does not rely on a fixed one-shot compression strategy. Rather, it continuously adjusts precision allocation according to the needs of different training stages. Compared with static quantization, this dynamic mechanism helps avoid over-compression at critical moments and thus improves training stability.

A closer examination of a single Transformer block further shows that \STREAM is component-aware. As shown in Figure~\ref{fig:block0}, LayerNorm-related parameters usually retain higher bit-widths, while many parameters in the MLP can be compressed more aggressively. This indicates that precision sensitivity varies not only across layers, but also across operator-level within the same layer.

Therefore, the strength of \STREAM is not merely that it lowers the average bit-width. Instead, it dynamically allocates precision across layers, stages, and components, so that the limited precision budget is used where it matters most. As a result, \STREAM can greatly reduce the memory overhead of optimizer states while preserving convergence behavior and representation quality close to full-precision AdamW.

%\vspace{-12pt}

\subsection{Fine-tuning}
To evaluate the effectiveness of \STREAM on downstream tasks, we further perform fine-tuning experiments on natural language, vision, and multimodal benchmarks, as summarized in Table~\ref{tab:stream_finetune}

\textbf{Natural language tasks.} \STREAM obtains an accuracy of 0.9032 on MNLI. This result is close to the 0.9060 achieved by full-precision AdamW and higher than the 0.9002 of 8-bit AdamW. On the wikitext-103 language modeling task, \STREAM reaches a perplexity of 20.1. This is also close to the full-precision baseline of 20.02 and better than the 20.22 of 8-bit AdamW. These results indicate that \STREAM can maintain stable optimization performance in both discriminative and generative language tasks.

\textbf{Vision tasks.} \STREAM achieves an mAP of 0.72434 on the COCO 2017 classification task. This result remains close to both full-precision AdamW, which obtains 0.72669, and 8-bit AdamW, which reaches 0.73082. Therefore, the dynamic quantization strategy of \STREAM also generalizes well to vision models, without causing obvious performance degradation.

\textbf{For multimodal tasks.} \STREAM achieves a Recall@1 of 0.7320 on the COCO 2017 image-text retrieval task. Not only is this result higher than the 0.7220 of 8-bit AdamW, but it also surpasses the 0.7240 of full-precision AdamW. This indicates that \STREAM can preserve stable optimization in unimodal tasks. Moreover, it can also support more complex multimodal representation learning.

In terms of memory overhead, fixed 8-bit AdamW consistently saves 75.0\% of optimizer-state memory across all tasks. In contrast, \STREAM further increases this saving to 82.5\%--87.5\%. Specifically, it achieves 84.4\%, 83.0\%, 82.5\%, and 87.5\% on MNLI, visual classification, language modeling, and image-text retrieval, respectively. Taken together, these results show that \STREAM can systematically surpass the compression limit of static 8-bit quantization through dynamic bit-width allocation.

%\vspace{-10pt}

\subsection{Ablation Study}
\begin{table}[t!]
\centering
\caption{Ablation study of \STREAM on WikiText-103 with GPT-2. Lower AvgBit and PPL indicate better compression and language modeling performance, respectively.}
\label{tab:stream_ablation}
\setlength{\tabcolsep}{7.5pt}
\renewcommand{\arraystretch}{1.05}
\small
\begin{tabular}{ccccc}
\toprule
Method & AvgBit $\downarrow$ & $\Delta$AvgBit & PPL $\downarrow$ & $\Delta$PPL \\
\midrule
\rowcolor{gray!20}
\STREAM (Full)      & 6.3 & 0.0  & 127.3 & 0.0  \\
w/o Dual Factor    & 6.4 & +0.1 & 130.2 & +2.9 \\
\rowcolor{gray!20}
w/o Temporal Factor& \textbf{5.8} & -0.5 & 128.7 & +1.4 \\
w/o Spatial Factor & 8.0 & +1.7 & \textbf{125.6} & -1.7 \\
\bottomrule
\end{tabular}
\end{table}
To analyze the role of each component in \STREAM, we conduct an ablation study on WikiText-103 with GPT-2. Table~\ref{tab:stream_ablation} reports the average AvgBit and PPL of different variants during the convergence stage, i.e., from steps 7000 to 8000.

As shown in Table~\ref{tab:stream_ablation}, removing the Dual Factor increases PPL by 2.9 compared with the full model, while AvgBit increases by only 0.1. This result indicates that the Dual Factor is the most critical component for preserving performance. In particular, the intensity and variation factors are important for characterizing the sensitivity of optimizer states. Therefore, removing this component causes clear performance degradation, even though the average bit-width changes only marginally.
In contrast, removing the Temporal Factor reduces AvgBit by 0.5, but increases PPL by 1.4. This indicates that temporal information is important for maintaining model stability under higher compression. Moreover, it helps the quantization process adapt better during the later stage of training.
Meanwhile, removing the Spatial Factor decreases PPL by 1.7, but increases AvgBit by 1.7. This result shows that the main role of the Spatial Factor is not to directly improve model performance. Instead, it mainly reduces the average bit-width and thereby improves overall compression efficiency.

Overall, these components play different but complementary roles in \STREAM. The Dual Factor mainly preserves performance, the Temporal Factor improves training-time quantization stability, and the Spatial Factor enhances compression efficiency. Therefore, although the full \STREAM is not the best variant on any single metric, it achieves a better balance between performance and compression ratio, which validates the effectiveness of the overall design.

% \vspace{-10pt}

\section{Conclusion}
In this paper, we propose \STREAM, a spatio-temporal aware dynamic quantization framework for optimizer states in large-scale multimodal model training. \STREAM jointly captures temporal training dynamics, spatial heterogeneity across layers, and gradient statistical features, and thereby enables adaptive bit-width allocation with very low overhead.

Experiments on language, vision, and multimodal tasks demonstrate that \STREAM achieves convergence stability and downstream performance comparable to, and in some cases better than, full-precision AdamW. Meanwhile, it reduces optimizer-state memory overhead by up to 84.4\% and compresses the average bit-width to 5.1 bits. Taken together, these results indicate that spatio-temporal aware dynamic quantization is an effective and general solution for optimizer compression in large-scale model training.

% We proposed \STREAM, a spatio-temporal aware dynamic quantization framework for optimizer states in large-scale multimodal training. By jointly modeling temporal dynamics, layer-wise heterogeneity, and gradient statistics, \STREAM enables low-overhead adaptive bit-width allocation. Across language, vision, and multimodal tasks, \STREAM matches or surpasses full-precision AdamW in convergence stability and downstream performance, while reducing optimizer-state memory overhead by up to 84.4\% and the average bit-width to 5.1 bits. These results validate spatio-temporal aware dynamic quantization as an effective and general approach to optimizer compression in large-scale training.

% \clearpage % 强制另起一页
\bibliographystyle{ACM-Reference-Format} % 这是ACM官方要求的格式，比unsrt更符合规范
\bibliography{references} % 你的文件名

\end{document}